\documentclass[journal]{IEEEtran}

\makeatletter
\long\def\@makecaption#1#2{\ifx\@captype\@IEEEtablestring%
\footnotesize\begin{center}{\normalfont\footnotesize #1}\\
{\normalfont\footnotesize\scshape #2}\end{center}%
\@IEEEtablecaptionsepspace
\else
\@IEEEfigurecaptionsepspace
\setbox\@tempboxa\hbox{\normalfont\footnotesize {#1.}~~ #2}%
\ifdim \wd\@tempboxa >\hsize%
\setbox\@tempboxa\hbox{\normalfont\footnotesize {#1.}~~ }%
\parbox[t]{\hsize}{\normalfont\footnotesize \noindent\unhbox\@tempboxa#2}%
\else
\hbox to\hsize{\normalfont\footnotesize\hfil\box\@tempboxa\hfil}\fi\fi}
\makeatother
\usepackage{stfloats}  
\usepackage{cite}
\usepackage{graphicx}
\usepackage{amsmath}
\usepackage{url}
\usepackage{multirow}
\usepackage{amssymb}
\usepackage{booktabs}
\usepackage{color}
\usepackage{soul}
\usepackage{array}
\usepackage[T1]{fontenc}

\ifCLASSINFOpdf
\else
\fi

\begin{document}
\bstctlcite{IEEEexample:BSTcontrol}

\title{Hate Speech Detection via Dual Contrastive Learning}

\author{Junyu~Lu,
		Hongfei~Lin,
		Xiaokun~Zhang,
		Zhaoqing~Li,
		Tongyue~Zhang,
		Linlin~Zong,
		Fenglong Ma,
		and~Bo~Xu*
\thanks{Junyu Lu, Hongfei Lin, Xiaokun Zhang, Zhaoqing Li, Tongyue Zhang, and Bo Xu are with the school of computer science and technology, Dalian University of Technology, China. Linlin Zong is with the school of software, Dalian University of Technology, China. Fenglong Ma is with the college of information science and technology, Pennsylvania State University, USA.  Corresponding Author: Bo Xu, e-mail: xubo@dlut.edu.cn}}

\markboth{Journal of \LaTeX\ Class Files,~Vol.~14, No.~8, August~2015}%
{Shell \MakeLowercase{\textit{et al.}}: Bare Demo of IEEEtran.cls for IEEE Journals}

\maketitle

\begin{abstract}

The fast spread of hate speech on social media impacts the Internet environment and our society by increasing prejudice and hurting people. Detecting hate speech has aroused broad attention in the field of natural language processing. Although hate speech detection has been addressed in recent work, this task still faces two inherent unsolved challenges. The first challenge lies in the complex semantic information conveyed in hate speech, particularly the interference of insulting words in hate speech detection. The second challenge is the imbalanced distribution of hate speech and non-hate speech, which may significantly deteriorate the performance of models. To tackle these challenges, we propose a novel dual contrastive learning (DCL) framework for hate speech detection. Our framework jointly optimizes the self-supervised and the supervised contrastive learning loss for capturing span-level information beyond the token-level emotional semantics used in existing models, particularly detecting speech containing abusive and insulting words. Moreover, we integrate the focal loss into the dual contrastive learning framework to alleviate the problem of data imbalance. We conduct experiments on two publicly available English datasets, and experimental results show that the proposed model outperforms the state-of-the-art models and precisely detects hate speeches. 
\end{abstract}


\IEEEpeerreviewmaketitle

\section{Introduction}

\IEEEPARstart{T}{he} widespread use of social media provides people with a broader space for communication and information exchange. People can freely express themselves on social media. While accelerating the dissemination of public opinions, social media also leads to the dissemination of undesirable speech, such as online hate speech. Nockleyby \cite{nockleyby2000hate} described hate speech as any kind of communication in speech, writing, or behavior, that attacks or uses pejorative or discriminatory language concerning a person or a group based on their religion, nationality, race, gender, or other identity factors. 

The ever-growing increase of online hate speech has become a pressing issue disturbing not only the groups which are humiliated and vilified but also the whole society due to the potential hate crimes \cite{Williams2019}. Even at the risk of restricting the freedom of expression, some social platforms have taken action against the proliferation of hate speech in ways of sealing accounts and removing content. 

The increasing social issue caused by online hate speech has attracted considerable attention of researchers in the natural language processing (NLP) field, seeking efficient and appropriate solutions to detecting online hate speech \cite{ding2019ynu_dyx, DBLP:conf/cikm/MouYL20, DBLP:conf/coling/CaoL20, DBLP:conf/acl/TekirougluCG20, zhou2021hate}. As early attempts to detect online hate speech, Chen \cite{DBLP:conf/socialcom/ChenZZX12} proposed lexical syntactic features to distinguish whether a sentence is hate speech. Mehdad \cite{DBLP:conf/sigdial/MehdadT16} detected hate speech using support vector machines (SVM) with sentiment features of a text. 

The state-of-the-art work has incorporated sentiment information for hate speech detection. Zhou et al. \cite{zhou2021hate} proposed the sentiment knowledge sharing (SKS) model integrated with an insulting word list and multi-task learning to detect hate speech. Although achieving promising performance in this task, the SKS model holds a strong assumption that \emph{insulting and negative emotions can distinguish between hate speech and non-hate speech}. However, this assumption cannot be always true as both hate and non-hate speeches may contain large amounts of negative words. Therefore, the SKS model with an insulting word list of hate speech achieved limited performance by overly focusing on the token-level emotional semantics. To further explain this phenomenon, we provide two example sentences from the SemEval-2019 Task-5 dataset \cite{basile2019semeval}, a publicly available dataset for hate speech detection.

\begin{center}
\begin{minipage}[h]{0.46\textwidth}
    \fbox
    {\small
    \parbox{\textwidth}{
    \textbf{Exp. 1} \textit{I can be a \underline{bitch} and an \underline{asshole} but I will love you and care about you more than any other person you have met. } (\textbf{Non-hate speech})\\
    \textbf{Exp. 2} \textit{Stop w ’we have to worry about the children’ No we do not-many R >20yrs old Go home and make your country better or enter ours legally we can’t afford them\#NODACA} (\textbf{Hate speech})
    }
    }
    \end{minipage}
\end{center}

It can be observed that although containing two insulting words, ``bitch'' and ``asshole'', the sentence in Exp. 1 is a non-hate speech as no attack is launched towards any social group. In contrast, Exp. 2 is a hate speech without any obvious abusive emotions, because it involves stereotypes of immigrant children. These two examples indicate that hate speech contains more complicated semantics and irregular expression patterns beyond negative emotions. 

To precisely detect hate speech, compared with the lexical sentiment, the trained models should focus on contextual semantic information to avoid the misclassification of non-hate speech containing abusive and insulting words. For a sentence with abusive or insulting words, the sentence does not contain hate speech if it is not targeted at certain social groups. According to the statistics in \tablename~\ref{example}, the speeches with insulting words account for a considerable proportion of two widely used hate speech detection datasets, SemEval-2019 Task-5 and Davidson et al. \cite{davidson2017automated}. However, \emph{effectively detecting these speeches with insulting words remains an unsolved problem in hate speech detection.}

Moreover, the datasets for hate speech detection mostly suffer from the problem of \textbf{data imbalance}. The imbalanced distribution of hate speech and non-hate speech would easily cause the detection model to pay too much attention to the class of non-hate speech with more samples, and ignore the class of hate speech with fewer samples, resulting in an imbalanced performance on data classification. Most existing methods are designed to optimize the overall performance, partly ignoring the data imbalance problem for hate speech detection.

\begin{table}
\renewcommand{\arraystretch}{1.2}
\centering
\caption{Proportion of samples containing insulting words on SemEval-2019 Task-5 and Davidson datasets.}
\begin{tabular}{{m{2cm}m{1.25cm}m{1.25cm}}}
\hline
\multicolumn{3}{l}{The SemEval-2019 Task-5 dataset}  \\
\hline
Label & \#Samples & Proportion\\
\hline
Hate speech & 2812 & 55.85\% \\
Non-hate speech & 2730 & 39.36\% \\
\hline
\hline
\multicolumn{2}{l}{The Davidson dataset} \\
\hline
Label & \#Samples & Proportion\\
\hline
Hate speech & 1147 & 80.21\% \\
Non-hate speech & 19756 & 84.60\% \\
\hline
\end{tabular}
\label{example}
\end{table}

To solve the above-mentioned problems, we propose a novel dual contrastive learning (DCL) framework for hate speech detection, 
which is tailored for the hate speech detection task by comprehensively considering the task-specific features, such as the subjectivity and contextualization of hate speech \cite{ross2016measuring}. Specifically, our model integrates both self-supervised and supervised contrastive learning, enriching the semantic representations of hate speech with context information itself and supervised signals from labels, effectively mitigating the misclassification of non-hate speech containing abusive and insulting words. 
Furthermore, since self-supervised contrastive learning has stronger adaptability than supervised contrastive learning from labels \cite{anand2020contrastive}, the representations learned from self-supervised contrastive learning can be considered as prior knowledge, facilitating the supervised classifications of hate speech by our DCL model. Therefore, we design the self-supervised contrastive learning before the supervised contrastive learning in DCL.
In addition, we introduce the focal loss, a reshaped cross entropy loss, to alleviate the problem of data imbalance. The main contributions of this work are summarized as follows.

\begin{itemize}
    \item We propose a dual contrastive learning framework for hate speech detection, particularly addressing the detection of hate speech containing insulting words by mining context information of data beyond the token-level emotional semantics.
    \item We integrate self-supervised and supervised contrastive learning into the focal loss to tackle the problem of data imbalance in hate speech detection.
    \item We examine the effectiveness of our model on two publicly used hate speech detection datasets, and demonstrate that our model can achieve state-of-the-art performance compared with the baseline models.
\end{itemize}

\section{Related Work}
We discuss two categories of related work: hate speech detection methods and contrastive learning methods.

\subsection{Hate Speech Detection Methods}
Detecting hate speech is a challenging natural language processing (NLP) task. 
Early work has used machine learning methods in automatically detecting hate speech. Davidson et al. \cite{davidson2017automated} presented a large-scale dataset and used Logistic Regression \cite{wright1995logistic} and SVM \cite{joachims1998making} with effective n-gram features for hate speech detection. These machine learning based methods can obtain the token-level features but mostly ignore the contextual semantic information that is highly needed for precise detection of hate speech, leading to limited performance.    

In recent years, with the development of deep learning and large-scale pre-training language models, many advanced models were proposed and achieved outstanding performance in hate speech detection. Several researchers use word embeddings obtained from unsupervised training on a large number of corpora to detect hate speech. Ding et al. \cite{ding2019ynu_dyx} used the FastText \cite{bojanowski2017enriching} tools to acquire word representations and presented a stacked Bidirectional Gated Recurrent Units (BiGRUs). Mou et al. \cite{DBLP:conf/cikm/MouYL20} proved the effectiveness of FastText and BERT \cite{devlin2019bert} for exploiting word-level semantic information and sub-word knowledge to identify hate speech. \cite{DBLP:conf/coling/CaoL20} proposed a reinforcement learning model HateGAN to address the problem of imbalance class by data augmentation. \cite{DBLP:conf/acl/TekirougluCG20} presented a hate speech detection dataset and used GPT-2 \cite{radford2019language} to pre-train the detection model. \cite{zhou2021hate} proposed the sentiment knowledge sharing (SKS) model combined with a negative word list and multi-task learning for hate speech detection. \cite{alkhamissi-etal-2022-token} evaluated the effectiveness of model to introduce infusing knowledge on out-of-distribution data. 
\cite{DBLP:conf/emnlp/ElSheriefZMASCY21, DBLP:conf/acl/HartvigsenGPSRK22, lu2023facilitating} facilitated the detection of implicit hate speech.
Previous research shows that deep learning based models can better obtain contextual information.
In addition, compared with the normative data in NLI tasks, hate speech crawled from social media is more nuanced, subjective, and contextual \cite{ross2016measuring}, which presents a huge challenge to natural language understanding. It is imperative to consider task-specific characteristics, such as the subjectivity and contextualization of hate speech, in designing effective detection models. Moreover, previous research has also demonstrated that the general methods of NLI task have limited performance in hate speech detection task \cite{fortuna-etal-2022-directions}. Therefore, we propose a dual contrastive learning method for hate speech detection.

\subsection{Contrastive Learning Methods}

Contrastive learning learns representations by contrasting positive and negative samples \cite{gutmann2012noise} and it has been widely employed in computer vision tasks \cite{DBLP:conf/eccv/GansbekeVGPG20, DBLP:journals/ijcv/LiYPLHP22, DBLP:journals/pami/LinGLBLP23, DBLP:journals/pami/HuZLPZP23, DBLP:journals/pami/Yang00B0023, DBLP:conf/aaai/Li0LPZ021, DBLP:journals/jmlr/PengLTZLZ22, DBLP:journals/tip/PengFXYZY18} for extracting in-depth supervision signals from the data itself. Nan et al. \cite{Nan_2021_CVPR} introduced a dual contrastive learning approach to better align text and video. Han et al. \cite{Han_2021_CVPR} proposed a novel method based on contrastive learning and a dual learning setting (exploiting two encoders) to infer an efficient mapping between unpaired data. Li et al. \cite{li2021contrastive} proposed a contrastive learning framework to learn instance and cluster representations. 

Contrastive learning has a wide range of applications in NLP, seeking for learning high-dimensional latent features of sentences by reducing reconstruction error. For example, Gao et al. \cite{gao2021simcse} used standard dropout as noise twice for a sentence embedding to build contrastive samples and proposed SimCSE to calculate semantic similarity. \cite{gunel2020supervised} and \cite{moukafih:hal-03367972} proposed supervised contrastive loss combined with cross-entropy to train a classification model for natural language understanding. \cite{DBLP:journals/ijcv/LiYPLHP22} proposed a self-supervised clustering with contrastive learning for general NLI tasks. This method integrates both instance-level and cluster-level self-supervised contrastive learning to obtain pseudo labels, which are further used for representation learning. However, due to the subjectivity and contextualization of hate speech \cite{ross2016measuring}, pseudo labels generated by general self-supervised methods would become unreliable and difficult to use to determine whether a sentence contains hate speech.
Totally different from \cite{DBLP:journals/ijcv/LiYPLHP22}, we propose a dual contrastive learning method for the task of hate speech detection. By considering the task-specific characteristics shown in Section II.A, our model integrates both self-supervised and supervised contrastive learning to enrich the semantic representations of hate speech.


\begin{figure*}
\centering
\includegraphics[width=16cm]{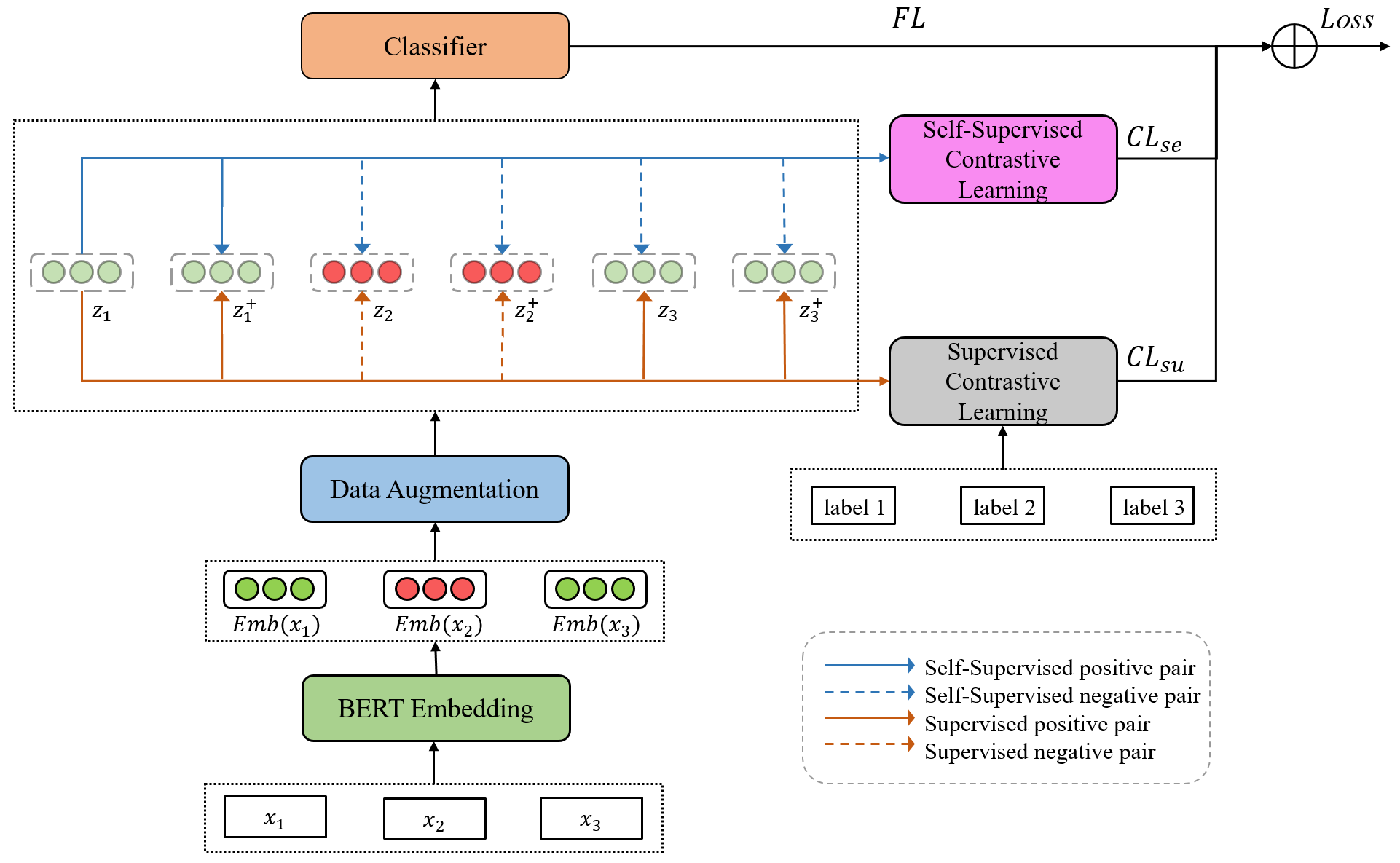}
\caption{The overall framework of our model. $CL_{se}$ and $CL_{su}$ are short for the self-supervised contrastive loss and the supervised contrastive loss, respectively. $FL$ represents the focal loss. $Loss$ represents the final loss function. The colors of circles denote the labels of sentences, embedded as $Emb(x_i)$. Based on $Emb(x_i)$, two augmented samples $z_j$ and $z_{j}^+$ can be generated using independently sampled dropout masks. Given $z_j$ as the reference object, the solid blue/orange arrows point to the positive samples of $z_j$ in $CL_{se} / CL_{su}$, while the dashed blue/orange arrows point to the contrastive samples in $CL_{se} / CL_{su}$.}
\label{model}
\end{figure*}

\section{Methodology} \label{sec_approach}
In this section, we introduce our model named DCL for hate speech detection. Our model seeks to learn adversarial samples using dual contrastive learning mechanisms. We first illustrate the overall framework of our model and then introduce the self-supervised contrastive learning and the supervised contrastive learning used in our model. Besides, we provide more implementation details for easily reproducing our model. 

\subsection{Overall Framework}
\figurename~\ref{model} shows the overall framework of our DCL model for hate speech detection. The input of our framework is a set of sentences including hate and non-hate speeches. Pre-trained BERT \cite{devlin2019bert} is employed to represent the sentences, and data augmentation is performed for two-stage contrastive learning. The first stage adopts self-supervised contrastive learning to make our model learn representations that are invariant to different views of positive pairs of hate speech, which are generated from the same sample by strong data augmentation, while maximizing the distance between negative pairs of non-hate speech. In the second stage, supervised contrastive learning utilizes the label information to pull clusters of points belonging to the same class together in embedding space, while pushing apart clusters of samples from different classes. Finally, we integrate the dual contrastive learning objectives into the focal loss for model optimization to alleviate the problem of data imbalance in hate speech detection.

\subsection{Self-Supervised Contrastive Learning}
Considering the complicated expressions and ambiguous semantics in hate speech expressions, we propose to use self-supervised contrastive learning for data augmentation and deep semantic information mining. By building positive and negative samples, self-supervised contrastive learning captures more comprehensive span-level features beyond token-level semantics for effectively distinguishing different samples. For hate speech detection, we propose a self-supervised contrastive learning method for mining potential useful semantic information of sentences in the model training process. 

Our self-supervised contrastive objective intends to distinguish positive samples constructed by data augmentation for each input sample against a set of negative samples in each batch of data. Inspired by a simple yet powerful sampling strategy \cite{gao2021simcse}, we predict the input sentences itself with dropout noises \cite{DBLP:journals/jmlr/SrivastavaHKSS14} to retain the maximum semantic information of hate speech. Other sampling strategies can also be integrated in our framework, which remains as future work. 

Specifically, for an input sentence $x_i$, we use standard dropout as noise twice for each sentence embedding, denoted as $Emb(x_i)$. Based on $Emb(x_i)$, two augmented samples $z_j$ and $z_{j}^+$ with respect to $x_i$ can be generated using independently sampled dropout masks placed on fully-connected layers. $(z_j, z_j^+)$ is regarded as a pair of positive samples, and other samples in the same batch are treated as negative ones. Based on this idea, our self-supervised contrastive learning loss for hate speech detection can be formulated as follows.
\begin{equation}
CL_{se} = -\sum_{j=1}^{2N}log \frac{e^{sim(z_{j}, z_j^+)/\tau_{se}}}{\sum_{k=1}^{2N}1_{[j\neq k]}\cdot e^{sim(z_{j}, z_{k})/\tau_{se}}}
\end{equation}
where $N$ denotes the batch size before data augmentation and $\tau_{se}$ is a non-negative temperature hyperparameter. $sim(\cdot)$ is the similarity scoring function between $z_j$ and $z_{j}^+$. In our implementation, we adopt the cosine similarity to capture the contextual semantic information by reconstructing the input samples, namely, $sim(z_j ,z_j^+) = \frac{{z_{j}}^Tz_{j}^+}{\left\|{z_{j}}\right\| \left\|{z_{j}^+}\right\| }$. 

\subsection{Supervised Contrastive Loss}
Self-supervised contrastive learning augments the training data by highlighting the Span-level semantics of hate speech from the data itself. To further incorporate supervised signals for hate speech detection, we use supervised contrastive learning on the basis of the augmented data. In other words, our supervised contrastive learning method integrates label information into the embedding space of the input sentences. The learned sentence embedding contrasts a set of positive samples against a set of negative samples in the same batch. Compared with self-supervised contrastive learning,  supervised contrastive learning incorporates more supervised information by considering more positive samples for each sampling batch. Specifically, for a batch of data with $N$ samples, supervised contrastive loss can be formulated as follows: 
\begin{equation}
\begin{aligned}
CL_{su} =& -\sum_{i=1}^{N}\frac{1}{N_{y_i}-1}\sum_{j=1}^{N} 1_{[i\neq j]} \cdot 1_{[y_i= y_j]}\\
& \cdot log \frac{e^{sim(z_i, z_j)/\tau_{su}}}{\sum_{k=1}^{N}1_{[i\neq k]}\cdot e^{sim(z_i, z_k)/\tau_{su}}}
\end{aligned}
\end{equation}
where ($z_i, z_j)$ denotes a pair of positive samples, and $(z_i, z_k)$ denotes a pair of randomly selected samples. $y_{i}$ and $y_j$ denotes the label of $z_i$ and $z_j$, respectively. $N_{y_i}$ is the number of samples with the same label as $z_{i}$. $\tau_{su}$ is the non-negative temperature coefficient of supervised contrastive loss. $CL_{su}$ further guides the model with supervised information for building effective detection models. To jointly combine self-supervised and supervised information, we use an overall loss function of contrastive learning as follows: 
\begin{equation}
\begin{aligned}
CL=CL_{se} +CL_{su}.
\end{aligned}
\end{equation}

\subsection{DCL Integrating Focal Loss}
We represent the input sentences using the pre-trained language model BERT \cite{devlin2019bert}. Any sentence $x_i$ is embedded as representations denoted as $Emb(x_i) \in R^{n\times d_{emb}}$, where $n$ is the sequence length of $x_i$, and $d_{emb}$ is the dimension of the embedding. A max-pooling layer is then applied to convert $Emb(x_i)$ into a vector representation $z_i \in R^{1\times d_{emb}}$ that is treated as the sentence embedding of $x_i$. Given $z_i$, we can predict the target class of $x_i$ using the softmax function:
\begin{equation}
    p(c|z_i) = softmax(z_i W)
\end{equation}
where $W \in d^{dim \times N_c}$ is a learnable parameter matrix. $c$ is the target class of $x_i$. $N_c$ is the number of classes.  
To estimate the inconsistency between the predicted label and the target label, we adopt the focal loss \cite{lin2017focal} that has been confirmed effective in imbalanced data classification. Since hate speech detection suffers from the problem of data imbalance, we introduce the focal loss to reshape the standard cross entropy loss such that the loss assigned to well-classified samples receives lower weights. The focal loss for hate speech detection is defined as:
\begin{equation}\label{eq:FL}
    FL = -\sum_{i=1}^{N}{\alpha_i(1-\hat{p_i})^\gamma log(\hat{p_i})}
\end{equation}
where $\gamma$ is a non-negative tunable focusing parameter to differentiate between easy and difficult samples. A smaller value of $\gamma$ guides the learned model to focus more on the misclassified samples, and meanwhile reducing the relative loss for well-classified samples. $\alpha \in [0, 1]$ is a weighting factor to balance the importance of positive and negative samples, which is defined as: 
\begin{equation}
    \alpha_i=\left\{
\begin{aligned}
\alpha && if \quad y_i = 1 \\
1-\alpha && otherwise 
\end{aligned}
\right.
\end{equation}

$\hat{p_i}$ in Eq.~(\ref{eq:FL}) reflects the relationship between the estimated probability and the target class.
\begin{equation}
\hat{p_i}=\left\{
\begin{aligned}
p_i && if \quad y_i = 1 \\
1-p_i && otherwise 
\end{aligned}
\right.
\end{equation}
where $p_i\in [0, 1]$ is the estimated probability for the class with the label $y_i=1$ in each sentence embedding $z_i$. 
During the training phase, the focal loss and the contrastive learning loss are jointly optimized. To learn a more robust model, we introduce a weighting coefficient $\lambda$ to balance the impact of these two loss functions. The final loss is defined as: 
\begin{equation}
Loss = FL + \lambda \cdot CL,
\end{equation}
where $\lambda \in [0, 1]$ is the weighting coefficient.  

\section{Experiments}
In this section, we evaluate the performance of our model. We first introduce the two commonly-used datasets, experimental settings, and baselines, and then present the evaluation results of our model compared with other baseline models. 

\subsection{Datasets}
We conduct our experiment on two publicly available datasets, which have been widely used in related research for comparison of hate speech detection models. The details of these datasets are introduced as follows:

\textbf{SemEval-2019 Task-5 (SE)} The SE dataset came from the Task-5 of SemEval-2019 \cite{basile2019semeval}. The subtask A of this evaluation is hate speech detection. The hate speech of this dataset is against women and immigrants. The total number of data is 11,971, where 5,035 data are labeled as hate speech, and the remaining 6,936 data belong to the non-hate class. This dataset contains three subsets: The training set with 9000 samples, the validation set with 1000 samples, and the test set with 2971 samples.

\textbf{Davidson Dataset (DV)} The DV dataset was constructed by Davidson et al. \cite{davidson2017automated}. The data were collected from tweets that contained hate speech including racist, sexist, homophobic, and offensive expressions in various ways. This dataset consists of 24,783 tweets, where only 1,430 ones are labeled as hate, and 23,353 data are non-hate. We can observe that this dataset is an extremely imbalanced dataset with relatively very few positive samples of hate speech.

\subsection{Experimental Settings}
We use BERT for representing the input sentences, which is fine-tuned on the downstream detection tasks. The pooling layer of bert-base-cased is taken as 768-dimensional sentence embedding. We use the 0.5 dropout rate and the AdamW optimizer \cite{loshchilov2018fixing} for model training. The learning rate is set to be 1e-4 and the batch size as 128. We set $\tau_{se}=0.1$ in the self-supervised contrastive loss, $\tau_{su}=0.05$ in the supervised contrastive loss and $\alpha = 0.3$ and $\gamma = 2$ in the focal loss. All models were trained on NVIDIA GeForce GTX 1080 GPU.

To compare with baseline methods, we use accuracy (Acc) and F-measure (F1) as evaluation metrics and import the experimental results of baseline methods from the literature. Since the SE dataset is from an evaluation task, the reported experimental results are based on the performance of the test set of the official evaluation. We select the models and hyperparameters that perform best on the validation set and evaluate the performance on the test set. Results are evaluated based on the officially designated metrics, including accuracy (Acc) and macro F1. For the DV dataset, we adopt the mean accuracy and the weighted F1 after five-fold cross-validation, and save the parameters corresponding to the optimal model, which follows the settings in previous work \cite{zhou2021hate}. 
We used the different F1 score metrics on two datasets following existing studies, such as the SOTA baseline model SKS \cite{zhou2021hate} for fair comparisons. 
In fact, the macro-F1 metric used for the SE dataset is a common choice in related tasks, while the weighted-F1 metric is a tailored version of macro-F1 for the DV dataset by considering that the DV dataset is very unbalanced with a ratio of hate to non-hate of about 1:15. If macro-F1 is used on DV, the performance of hate samples will dominate the overall performance. Therefore, to make more reasonable evaluations of different models on DV, weighted F1 is designed for this dataset, which considers the weights of hate and non-hate samples.

\subsection{Baseline Methods}
We compare our model with the following baselines:

\textbf{SVM.} The SVM-based hate speech detection model was proposed by Zhang et al. \cite{zhang2018detecting} and Mandl et al. \cite{mandl2019overview}. The researchers extracted several statistical features, such as n-gram, insulting words, and the frequency of particular punctuation marks for learning SVM classifiers.

\textbf{LSTM, GRU, Bi-LSTM.} These methods were proposed by Ding et al. \cite{ding2019ynu_dyx}. They employed word embedding and learned sentence representations using LSTM, GRU, and Bi-LSTM to detect hate speech, respectively.

\textbf{CNN-GRU.} Zhang et al. \cite{zhang2018detecting} applied convolution-GRU based deep neural network with word embedding to extract potential semantic features in detecting hate speech, which captures both word sequential and order information in tweets. 

\textbf{BERT.} This baseline was proposed by Benballa et al. \cite{benballa2019saagie}. The final hidden state of [CLS] of BERT is used as the sentence embedding in hate speech detection. The classifier consists of a feed-forward layer and a softmax function. For a fair comparison, we train the model using cross-entropy loss and focal loss, respectively. 

\textbf{SKS.} It was proposed by Zhou et al. \cite{zhou2021hate}. This approach detected hate speech based on sentiment knowledge sharing and achieved state-of-the-art performance on the Davidson dataset and SemEval-2019 Task-5, which is a strong baseline for comparison.

\begin{table}[!t]
\renewcommand{\arraystretch}{1.2}
\centering
\caption{Comparison with baselines on SE and DV. The results with an asterisk (*) are imported from the literature.}
\begin{tabular}{l|cc|cc}
\hline
Dataset & \multicolumn{2}{c}{SE} & \multicolumn{2}{|c}{DV} \\
\hline
Metrics & Acc. & macro-F1 & Acc. & weighted-F1\\
\hline
SVM* & 49.2 & 45.1 & - & 87.0 \\
LSTM* & 55.0 & 53.0 & 94.5 & 93.7 \\
GRU* & 54.0 & 52.0 & 94.5 & 93.9 \\
BiLSTM* & 53.5 & 51.9 & 94.4 & 93.7 \\
CNN-GRU* & 62.0 & 61.5 & - & 94.0 \\
BERT(BCE) & 55.8 & 54.9 & 94.3 & 94.2 \\
BERT(FL) &59.8  &58.6  &94.4  &94.4  \\
SKS* & 65.9 & 65.2 & 95.1 & \textbf{96.3} \\
\hline
DCL (R) & 65.9 & 63.1 & 94.8 & 94.7\\
DCL (Ours) & \textbf{67.8} & \textbf{67.2} & \textbf{95.9} & 95.6 \\
\hline
\end{tabular}
\label{mainresults}
\end{table}

\subsection{Results and Discussions}
\tablename~\ref{mainresults} shows our evaluation results on the SE and DV datasets. From \tablename~\ref{mainresults}, we can observe that:

(1) Overall, the experimental performance on these two datasets is largely different. On the DV dataset, the values of the two used metrics are both above 93\%. While on the SE dataset, the values are less than 70\%. This is because the data distributions of these datasets differ a lot. Namely, subtle differences in data distributions can significantly affect the detection performance.

(2) The performance of neural network-based models is much better than the SVM-based models with manually crafted features. Compared with LSTM and its variants, hybrid neural networks, such as CNN-GRU achieved better performance, particularly on the SE dataset. Furthermore, SKS, benefiting from its sentiment knowledge-sharing mechanism and multi-task learning, achieved the best performance among all the baselines.

(3) Our model DCL outperformed all the baseline models on the SE dataset. The improvement of DCL over the BERT-based model is 13\%, and the improvement over LSTM, GRU, and SVM is more 10\% in terms of the macro-F1 and the accuracy. Compared with the best-performed baseline model SKS, DCL is superior in terms of both metrics.

(4) On the DV dataset, DCL achieved the best performance by accuracy, and better performance by weighted-F1 than all the baseline models except SKS. This is because \textbf{SKS used an external sentiment dataset} to enhance the performance. Although DCL does not use any external data, DCL achieved higher accuracy than SKS. 

(5) We further analyze the impact of the sequence between the two stages. Specifically, we reverse the order of self-supervised and supervised contrastive learning, referred to as DCL(R). As the result shown in \tablename~\ref{mainresults}, regardless of the order of DCL, it has a more competitive performance than baselines on the two datasets. Meanwhile, if self-supervised contrastive learning is before supervised contrastive learning, DCL has better detection effects. This is because the features learned from self-supervised contrastive learning represent the context information of the text itself and they are more adaptive than supervised training \cite{anand2020contrastive}. They can be considered as prior knowledge facilitating model decisions on downstream tasks. Therefore, it is more reasonable to employ self-supervised comparative learning as the first stage of DCL. 

\begin{figure}
\centering
\includegraphics[width=8.5cm]{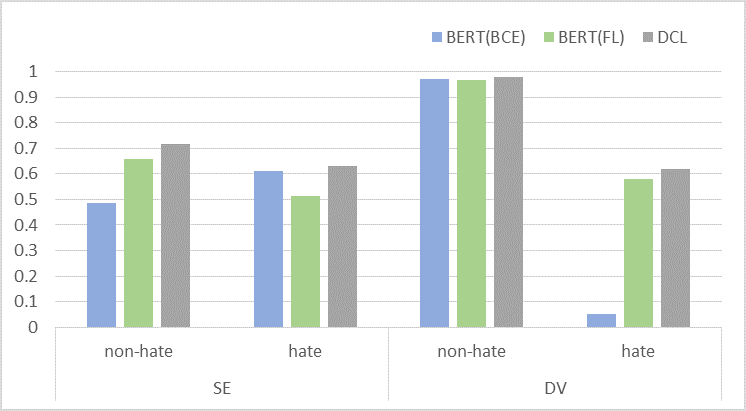}
\caption{F1-Score of hate and non-hate sentences on SE and DV. Blue: BERT trained with BCE, green: BERT trained with Focal loss, gray: DCL.}
\label{fig:fakehatespeech}
\end{figure}

(6) Figure \ref{fig:fakehatespeech} shows the F1-Score of detection performance of hate and non-hate samples on SE and DV. From the figure, we observe that our model has the more advanced performance to distinguish whether the sentences contain hate speech than BERT trained with binary cross entropy or focal loss. This result indicates that the use of focal loss integrated with dual contrastive learning largely alleviates the data imbalance problem of hate speech detection. For DV, we find that the capability of hate speech detection is much lower than that of non-hate speech on a model trained using only cross-entropy. This is because the DV dataset is extremely imbalanced, which partly hinders the improvement of model performance. 

To further validate the ability of dual contrastive learning in reconstructing text representation, we use t-Distributed Stochastic Neighbor Embedding (t-SNE) \cite{Maaten2008} to plot the learned sentence embedding $z_i$. t-SNE is utilized to reduce the dimension of representations from high-dimensional vector space to a two-dimensional plane. Since the number of hate speech on DV is fewer, we perform the t-SNE based plotting only on the test set of SE that contains 1180 hate speeches and 1625 non-hate speeches.

\begin{figure}
\centering
\includegraphics[width=8.75cm]{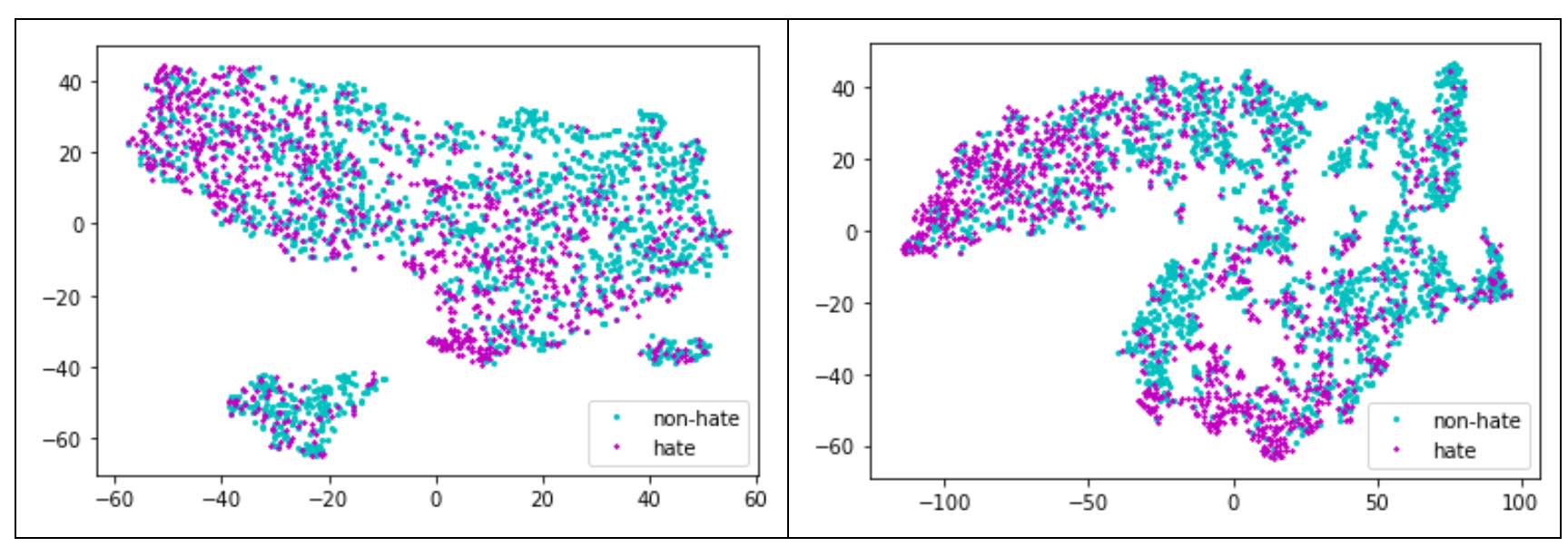}
\caption{t-SNE plots of the learned sentence-level embedding $z_i$ on SemEval-2019 Task-5 test set using the BERT model (left) and our model (right). Cyan: non-hate examples; Pink: hate examples.}
\label{fig:t-SNE}
\end{figure}

We illustrate the t-SNE plots of the learned sentence embeddings in \figurename~\ref{fig:t-SNE}. From the figure, we can observe that the distinction between hate speech and non-hate speech has been improved by introducing dual contrastive learning loss. Meanwhile, the vector space of the two classes still overlaps in certain dimensions, which indicates that some sentences with different labels have similar topical information such as immigrants. The vector representations of hate speech samples with the same topic tend to be closer than those with the same labels (i.e. hate and non-hate). This also leads to the fact that pseudo-labels generated by general self-supervised methods, such as \cite{DBLP:journals/ijcv/LiYPLHP22}, will become unreliable, making it difficult to determine whether a sentence contains hate speech or not.
To further investigate the effectiveness of the loss functions used in our model, we provide an ablation study in the next section.

\subsection{Ablation Experiments}
In this section, we investigate the influence of contrastive learning and the choice of weighting coefficient $\lambda$ in our model, respectively.

\subsubsection{The influence of contrastive learning.} \tablename~\ref{influenceCL} shows the influence of different parts of our model, where "-self" is the proposed model without the self-supervised contrastive learning, and "-sup" is the proposed model without supervised contrastive learning. 

\begin{table}
\renewcommand{\arraystretch}{1.2}
\centering
\caption{The result of ablation experiments.}

\begin{tabular}{l|cc|cc}
\hline
Dataset & \multicolumn{2}{c}{SE} & \multicolumn{2}{|c}{DV} \\
\hline
Metrics & Acc. & macro-F1 & Acc. & weighted-F1\\
\hline
-self & 64.4 & 63.0 & 95.1 & 95.0 \\
-sup & 57.5 & 57.2 & 95.8 & 95.5 \\
\hline
DCL & \textbf{67.8} & \textbf{67.2} & \textbf{95.9} & 95.6 \\
\hline
\end{tabular}

\label{influenceCL}
\end{table}

\begin{figure}
\centering
\includegraphics[width=8.5cm]{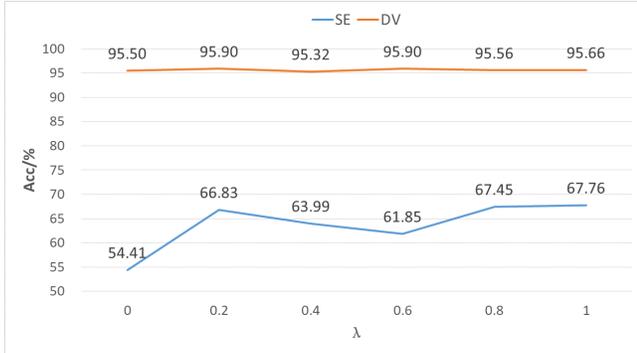}
\caption{The accuracy of the model under different $\lambda$. }
\label{lambda}
\end{figure}

Based on the results in \tablename~\ref{influenceCL}, we observe that: (1) The self-supervised contrastive learning loss contributes a lot on both datasets, which demonstrates that self-supervised contrastive learning can enhance the model's ability in acquiring the high-level semantic features of potentially hate speech. (2) On different datasets, the performance based on supervised contrastive learning is quite different. The performance decreases more sharply on SE than that on DV. The reason for this phenomenon is that the proportion of hate speech on DV is much lower than SE, and our model hardly obtained enough positive samples for supervised contrastive learning. On SE, samples are relatively balanced and supervised contrastive learning can make the best of positive and negative samples for learning an effective detection model. This finding indicates that the label information is significant to supervised contrastive learning in our model. 

\subsubsection{The choice of weighting coefficient \(\lambda\).} To further examine the influence of contrastive learning in DCL, we tune the weighting coefficient $\lambda$ and report the performance change in \figurename~\ref{lambda}. From the figure, we observe that on DV, the best performance of DCL can be achieved when $\lambda=0.2$ or $\lambda=0.6$, while on SE, the best performance is achieved when $\lambda=1.0$. The results indicate that contrastive learning exhibits higher performance on the balanced dataset SE, while the focal loss contributes more to the imbalanced dataset DV.

\subsection{Performance of Detecting the Speeches Containing Insulting Words}

In order to further verify whether our model has a stronger ability in detecting speech containing insulting words, we conducted this supplementary experiment. We first utilized an insulting vocabulary collected from Twitter\footnote{https://github.com/Mrezvan94/Harassment-Corpus} \cite{rezvan2018quality} and NoSwearing\footnote{https://www.noswearing.com/}, a website listing swear words. The vocabulary contains a total of 1060 frequently insulting words which are divided into six types of contexts: 1) Sexual 2) Appearance-related 3) Intellectual 4) Political 5) Racial 6) Combined. This resource is used to refine the samples with insulting words in the SE and DV datasets. The statistics of the refined datasets are illustrated in \tablename~\ref{example}, which indicates there is a large proportion of speeches containing insulting words in these two datasets. We then used the refined datasets to examine the detection performance of the learned model compared with the BERT-based model. The results on these refined datasets are reported in \tablename~\ref{fakehatespeech} and \figurename~\ref{fig:fakehatespeech}. 

\begin{table}
\renewcommand{\arraystretch}{1.2}
\caption{Performance of models trained on the samples containing insulting words.}
\centering
\begin{tabular}{l|cc|cc}
\hline
Dataset & \multicolumn{2}{c}{SE} & \multicolumn{2}{|c}{DV} \\
\hline
Metrics & Acc. & macro-F1 & Acc. & weighted-F1\\
\hline
BERT(BCE) & 64.4 & 63.0 & 98.3 & 98.4 \\
DCL & \textbf{70.6} & \textbf{70.1} & \textbf{98.8} & \textbf{98.8} \\
\hline
\end{tabular}
\label{fakehatespeech}
\end{table}

From \tablename~\ref{fakehatespeech}, we observe that the improvements on Acc. and macro-F1 are 6.2\% and 7.1\% on SE and 0.5\% and 0.4\% on DV, respectively. The experimental results showed that our model has a much stronger ability in detecting speeches containing insulting words than the BERT-based model. The dual contrastive learning and focal loss of our model unitedly contribute to the improved performance of hate speech detection.

\subsection{Detection Examples and Error Analysis}
\subsubsection{Detection Examples} One advantage of our model is its capability in capturing span-level features. In this section, we provide four case studies to illustrate this capability of our model compared with the BERT-base-cased detection model. \tablename~\ref{case} shows the detection results. From the table, we observe that our model can precisely detect these examples, but the BERT-based model wrongly predicts their labels. 

Although the first sentence has two negative words, ``\textit{threats}'' and ``\textit{lying}'', that express somewhat insulting emotions, the sentence is not an attack towards certain social groups. Therefore, this sentence does not contain hate speech. On the contrary, the second sentence, as an example of hate speech, does not contain any insulting words but involves a stereotype of immigrant children. Our model correctly predicts that it is hate speech, which demonstrates the effectiveness of our model. Similarly, the third and the fourth sentences both contain an abusive word, ``\textit{bitch}''. By considering the context of each sentence, only the fourth sentence expresses hatred. For text containing the same insulting words, our model can also make correct predictions. This is because our model learns more contextual semantic information by dual contrastive learning, which helps effectively distinguish different kinds of samples, particularly hate speech containing insulting words.

\begin{table*} 
\renewcommand{\arraystretch}{1.1}

\centering
\caption{Examples on hate speech detection of our model and BERT.}
\small
\begin{tabular}{m{1cm}<{\centering}|m{8cm}|m{1.25cm}<{\centering}|m{1.25cm}<{\centering}|m{1.25cm}<{\centering}} 
\hline
Index & \centering{Sentence} & Label & BERT & DCL \\ 
\hline 
1 & Like he ever kept out any threats. He's lying as usual. \#BuildThatWall & non-hate & hate & non-hate \\
\hline
2 & stop w 'we have to worry about the children' No we do not-many R\textgreater20yrs old Go home and make your country better or enter ours legally we can't afford them \#NODACA & hate & non-hate & hate \\
\hline
3 & I can be a bitch and an asshole but I will love you and care about you more than any other person you have met. & non-hate & hate & non-hate \\
\hline
4 & my own sister even called and said bitch you pregnant now hoe know I can't have another one & hate & non-hate & hate \\
\hline
\end{tabular}
\label{case}
\end{table*}

\begin{table*} 
\renewcommand{\arraystretch}{1.1}
\centering
\caption{Misclassified instances on hate speech detection of our model and BERT.}
\small
\begin{tabular}{m{1cm}<{\centering}|m{8cm}|m{1.25cm}<{\centering}|m{1.25cm}<{\centering}|m{1.25cm}<{\centering}} 
\hline
Index & \centering{Sentence} & Label & BERT & DCL \\ 
\hline
1 & >Harasses women and calls them bitch and crabby >They block him >Plays the victim & non-hate & hate & hate\\
\hline
2 & Bitch how is you gonna claim to be a "real" nigga, yet still on that hoe shit? I'm not understanding the math, i'm ???? & non-hate & hate & hate\\
\hline
3 & He is 100\% accurate. Diversity is our greatest weakness. Unity is our strength. United we stand, divided we fall. \#buildthewall \#deportthemall \#stoptheinvasion \#americafirst & hate & non-hate & non-hate\\
\hline
\end{tabular}
\label{error}
\end{table*}

\begin{figure}
\centering
\small
Exp. 1
\includegraphics[width=8.5cm]{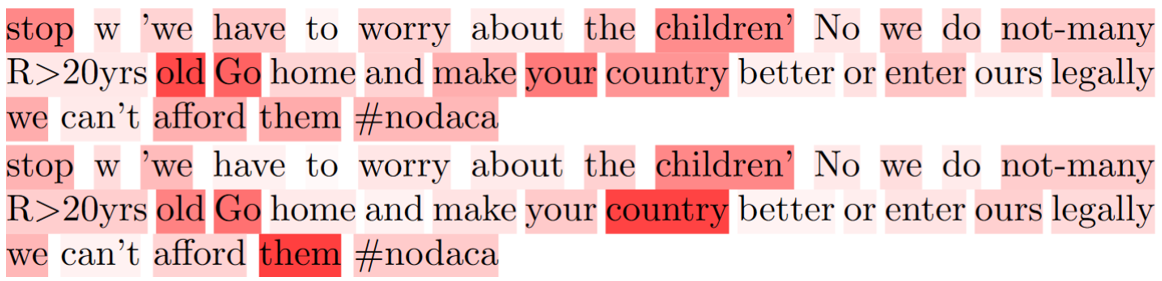}
Exp. 2
\includegraphics[width=8.5cm]{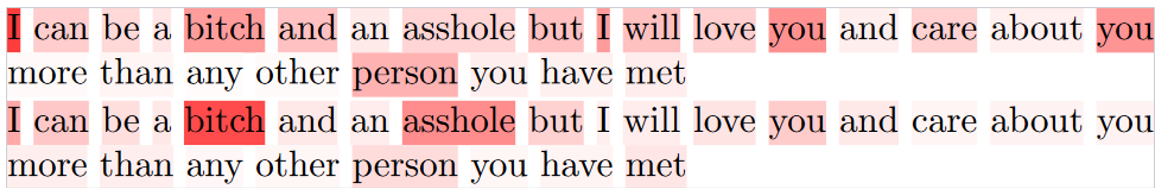}
\caption{Attention weights for each word of two sentences in the hidden layer of fine-tuned BERT encoder. For each sentence, the above one is trained with DCL, and the below one is trained with the BERT-based model. The depth of the background color indicates the weight of each word.}
\label{fig:attention}
\end{figure}

To further verify the effectiveness of our model, we visually analyze the attention weights of the hidden layers of fine-tuned BERT encoder in the learned DCL model and the BERT-based model through two sentences. The result is shown in \figurename~\ref{fig:attention}. For each sentence, the above one is trained with DCL and the below one is trained with the BERT-based model.

In \figurename~\ref{fig:attention}, the depth of red indicates the attention weight of the word. The darker the color, the more important the word is to the hate speech detection of the entire sentence. In Exp. 1, the word set \textit{\{Go, home, can't, afford\}} gets more attention from DCL than BERT. And in Exp. 2, the word set \textit{\{I, will, love, you\}} has a higher attention weight in sentences while the insulting words, such as ''\textit{bitch}'' and ''\textit{asshole}'', have a lower weight. The above sentences show that the model can better discover the key information of the context, which has a certain guiding significance for the hate speech detection task. 

\subsubsection{Error Analysis}
To gain more insights into the performance of our model, a manual inspection has been performed on a set of misclassified sentences. Two main types of error have been identified:

\textbf{Type I error} refers to the sentences annotated as \emph{non-hate}, but classified as \emph{hate} by the detection models. Type I error is usually caused by colloquial and informal statements in tweets. We enumerate two cases in \tablename~\ref{error} as examples: The first case describes the scene in an informal flowchart-like fashion, while the second case contains many colloquial languages, such as \textit{''gonna'', ''yet still''}, which is not conducive to the model's understanding of text semantics. Therefore, both models wrongly predicted their labels.

\textbf{Type II error} refers to the sentence labeled as \emph{hate}, but classified as \emph{non-hate} by the detection models. Type II errors usually occur when there is a lack of necessary background knowledge. For the third case in \tablename~\ref{error}, the meaning of this sentence is embodied by the information of hashtags, such as "\textit{\#buildthewall}", which reflect the hatred of opposition to racial diversity. Therefore, the stance contained in the hashtag needs to be considered as background knowledge in hate speech detection.

\section{Conclusion}
In this work, we propose a dual contrastive learning framework to tackle the problem of hate speech detection. Our framework integrates both self-supervised contrastive learning and supervised contrastive learning to capture high-level semantic information and complex language usage pattern in hate speech expressions. Furthermore, we integrate focal loss with dual contrastive learning to alleviate data imbalance for fine-grained hate speech detection. Experimental results on the SemEval-2019 Task-5 and Davidson dataset demonstrate the effectiveness of our model. 

In the future, we will explore the following directions: (1) The analysis of Type I error shows that noises in text affect the model's performance. Therefore, we will further explore the impact of insulting words in informal contexts on hate speech detection. (2) The analysis of Type II error certifies the necessity of external knowledge in hate speech detection. We will explore how to introduce useful external knowledge to further improve detection performance. 

\section{Acknowledgements}
This research is supported by the Natural Science  Foundation  of  China  (No. 62076046, 62006034), Natural Science Foundation of Liaoning Province (No. 2021-BS-067).

\ifCLASSOPTIONcaptionsoff
  \newpage
\fi

\bibliographystyle{IEEEtran}
\bibliography{custom}

\end{document}